\documentclass[11pt, a4paper, onecolumn]{lumia}

\usepackage[sort&compress]{natbib}
\setcitestyle{authoryear,round,citesep={;},aysep={,},yysep={;}}

\usepackage{graphicx}
\usepackage{subcaption}
\usepackage{booktabs}
\usepackage{colortbl}
\usepackage{xcolor}
\usepackage{makecell}
\usepackage{tabularx}
\usepackage{array}
\usepackage{mathtools}
\usepackage{nicefrac}
\usepackage{siunitx}
\usepackage{stfloats}
\usepackage{pdflscape}
\usepackage{rotating}
\usepackage{wrapfig}
\usepackage[export]{adjustbox}
\usepackage{enumitem}
\usepackage{algorithm}
\usepackage{algpseudocode}

\usepackage{url}
\usepackage{xurl}

\usepackage{placeins} 

\usepackage[capitalize,noabbrev]{cleveref}
\crefformat{section}{\S#2#1#3}
\crefformat{subsection}{\S#2#1#3}
\crefformat{subsubsection}{\S#2#1#3}

\usepackage{caption}
\usepackage[textsize=tiny]{todonotes}

\theoremstyle{plain}

\theoremstyle{definition}

\theoremstyle{remark}

\definecolor{color5}{HTML}{006795}
\hypersetup{
  colorlinks   = true,
  urlcolor     = color5,
  linkcolor    = color5,
  citecolor    = color5
}

\newlength{\colfigwidth}
\newcommand{\appendixref}[1]{\hyperref[#1]{Appendix~\ref*{#1}}}

\newcommand{\vect}[1]{\mathbf{#1}}

\setheadertext{LUMIA Lab}
\correspondingemail{\emailicon~boyizeng@sjtu.edu.cn, hyq718@sjtu.edu.cn, lin.zhouhan@gmail.com \quad $^\star$ Equal Contribution \quad $^\ddagger$ Corresponding Author.}
\renewcommand{\today}{2026-02-07}

\title{Pretraining with Token-Level Adaptive Latent Chain-of-Thought}

\author{%
    \textbf{Boyi Zeng}$^{1\star}$,
    \textbf{Yiqin Hao}$^{1\star}$,
    \textbf{He Li}$^{1}$,
    \textbf{Shixiang Song}$^{1,5}$,
    \textbf{Feichen Song}$^{1}$,
    \textbf{Zitong Wang}$^{1,4}$,
    \textbf{Siyuan Huang}$^{1}$,
    \textbf{Yi Xu}$^{3}$,
    \textbf{ZiWei He}$^{5}$,
    \textbf{Xinbing Wang}$^{3}$,
    \textbf{Zhouhan Lin}$^{1,2,5\ddagger}$\\
    \scriptsize
    $^1$ LUMIA Lab, School of Artificial Intelligence, Shanghai Jiao Tong University \\
    $^2$ Shanghai Al Laboratory \quad $^3$ Shanghai Jiao Tong University \quad $^4$ Sun Yat-sen University \quad $^5$ Shanghai Innovation Institute
}

\begin{document}

\begin{abstract}
Scaling large language models by increasing parameters and training data is increasingly constrained by limited high-quality corpora and rising communication costs. This work explores an alternative axis: increasing per-token computation without expanding parameters, by internalizing latent Chain-of-Thought (CoT) into pretraining. We propose \emph{Pretraining with Token-Level Adaptive Latent CoT} (adaptive latent CoT), where the model generates a variable-length latent CoT trajectory before emitting each token---allocating longer trajectories to difficult tokens and shorter (or even zero) trajectories to easy ones. Importantly, this behavior emerges naturally from one-stage pretraining on general text and reduces computation in both training and inference via token-wise adaptive halting. Experiments with Llama architectures show that adaptive latent CoT consistently improves language modeling perplexity and broad downstream accuracy, even with fewer training FLOPs than prior recurrent baselines.
\end{abstract}

\maketitle

\section{Introduction}
\label{sec:Introduction}

The remarkable success of Large Language Models (LLMs) has been primarily driven by scaling up model parameters and training data~\citep{kaplan2020scaling,liu2024deepseek,brown2020language}. However, this paradigm is approaching critical bottlenecks: the exhaustion of high-quality public data~\citep{muennighoff2023scaling,villalobos2022will} and the substantial communication overheads associated with scaling model size~\citep{pati2023computation,narayanan2021efficient,hoffmann2022training}. 
Consequently, a key research direction is to improve model capability under a fixed parameter and data budget by scaling compute per token.

To scale compute without expanding parameters, prior work has explored parameter-sharing strategies that emulate greater depth by recursively reusing the same layer weights~\citep{zeng2025pretraining,giannou2023looped,zhu2025scaling}. While appealing, this form of depth scaling can be prone to training instability. More recently, PonderLM2~\citep{zeng2025ponderlm} introduces latent computation before predicting the next token, but parallelizing this process often requires Jacobi-style iteration and incurs substantial training FLOPs. Moreover, these approaches typically allocate uniform compute to all tokens, which is inherently suboptimal compared to human cognition where \emph{thinking} is adaptive—brief for simple concepts and prolonged for complex reasoning.

Parallel to architectural modifications, another line of research has demonstrated that scaling inference-time compute—via CoT~\citep{wei2022chain}—has achieved remarkable success.  However, this capability typically relies on explicit supervision (annotated CoT data), is confined to a discrete token space, and is ultimately capped by the base model's pre-trained capabilities~\citep{yue2025does}.

These observations motivate a fundamental question: Can we internalize the benefits of Chain-of-Thought directly into pre-training, within a continuous latent space, and learn to allocate compute adaptively? In this work, we propose Pretraining LLMs with Adaptive Latent CoT, a pre-training framework that encourages the model to generate an adaptive sequence of latent CoT steps before emitting each observed token. The model learns to produce short (or even zero) latent trajectories for simple tokens while allocating longer latent chains for complex concepts. Unlike prior methods that require multi-stage training~\citep{fu2025think,zhu2025scaling,schuster2022confident}, additional supervision~\citep{elbayad2019depth}, only reduce inference compute~\citep{zhu2025scaling,schuster2022confident,banino2021pondernet}, or assume a prescribed halting prior~\citep{chen2025inner,raposo2024mixture,bae2025mixture,banino2021pondernet,zhu2025scaling}, our adaptive latent CoT emerges naturally from one-stage pre-training on general text, and can reduce computation in both training and inference.

\section{Related works}
\label{sec:related works}
\paragraph{Latent Reasoning}
To improve model capability without increasing parameters, an emerging line of work studies latent reasoning, where LLMs conduct reasoning internally in hidden states rather than via explicit verbalization.
One branch remains in the discrete vocabulary space by inserting specialized markers---e.g., \textsc{Pause}~\citep{goyal2023think}, \textsc{Filler}~\citep{pfau2024let}, or \textsc{Planning}~\citep{wang2023guiding} tokens---to allocate additional ``thinking time'' and induce extra implicit computation steps. Similarly, Quiet-STaR~\citep{zelikman2024quiet} uses reinforcement learning to generate token-level rationales. However, these approaches are fundamentally constrained by the discreteness of the vocabulary, which limits the expressiveness and flexibility of the reasoning process. Consequently, another line of work moves reasoning into continuous spaces. For example, Coconut and SIM-CoT~\citep{hao2024training,wei2025sim} leverage hidden states (e.g., the final-layer representations) as continuous inputs to simulate chain-of-thought. Despite encouraging results, most existing continuous latent reasoning methods still rely on supervised fine-tuning and curated CoT annotations, which hinders scalability.
\paragraph{Recursive Models}
In parallel, an orthogonal way to improve capability under a fixed parameter budget is recursive parameter sharing, which scales compute per token by iteratively reusing the same weights.
Universal Transformers~\citep{dehghani2018universal} show that weight sharing can provide strong capacity with minimal parameter growth, and ALBERT~\citep{lan2019albert} further improves efficiency via extensive sharing without hurting performance.
More recent work leverages recurrence to scale effective depth and performance: PonderLM2~\citep{zeng2025ponderlm} adds recursive latent computation before next-token prediction and reports strong gains, though parallelizing often relies on Jacobi-style iteration and can be FLOP-heavy.
Looped Transformers, PonderLM, and related methods~\citep{zeng2025pretraining,giannou2023looped,geiping2025scaling} similarly reuse parameters to increase depth and show benefits on general modeling and reasoning tasks.
However, as effective depth grows, training can become unstable, and compute is uniform across tokens regardless of input difficulty.
\paragraph{Adaptive Computation}
Many works adopt Adaptive Computation Time (ACT)~\citep{graves2016adaptive} to let models decide per token how many iterative steps to run, and this has inspired a broad class of adaptive token/layer routing methods.
For example, ITT~\citep{chen2025inner} uses a predefined per-layer token-selection ratio, while MoD/MoR~\citep{raposo2024mixture,bae2025mixture} activate only top-$k$ tokens.
PonderNet~\citep{banino2021pondernet} adds a geometric halting prior to stabilize learning.
However, these approaches typically fix or strongly constrain compute allocation a priori, limiting the model’s ability to explore adaptive policies during training. A complementary line achieves adaptivity via multi-stage training and inference-time early exiting, e.g., TaH~\citep{fu2025think}, Ouro~\citep{zhu2025scaling}, and CALM~\citep{schuster2022confident}.
Ouro, for instance, learns an exit gate with an entropy regularizer during pretraining and then freezes the LM while fine-tuning only the gate.
Other methods use extra supervision, such as Depth-adaptive Transformers~\citep{elbayad2019depth} that predict when to skip remaining layers using likelihood/correctness signals.
Moreover, methods like Ouro, PonderNet, and CALM~\citep{zhu2025scaling,banino2021pondernet,schuster2022confident} typically still require full computation during training, so they \emph{do not reduce training FLOPs} and mainly save compute at inference.

In contrast, we propose a one-stage pretraining framework that learns ACT end-to-end without a priori constraints, reducing computation in both training and inference.
\section{Method}
\label{sec:method}
We propose an efficient latent Chain-of-Thought (latent-CoT) training and inference framework that removes the sequential dependency bottleneck while attaining adaptive compute. 
Our method consists of three key components: 
(1) \emph{Parallel Masking} that extends attention causality to a 2D index $(t,k)$ (token position and latent step), enabling parallel computation over all positions at each latent step; 
(2) a \emph{probabilistic halting} mechanism with a lightweight Router that models continuation/exit probabilities, together with threshold pruning and mass-preserving mixing to obtain the final representation; and 
(3) a \emph{correctness-aware adaptive loss} that discourages unnecessary latent computation when the model is already confident on the ground-truth token.

\subsection{The Sequential Dependency Bottleneck}
Let $\vect{x} = \{x_1, \dots, x_L\}$ denote the input sequence.
In a standard latent CoT framework, for each position $t$ the model inserts a sequence of latent steps $\vect{z}_t = \{z_{t}^{(1)}, \dots, z_{t}^{(K)}\}$ before the subsequent target token $x_{t+1}$.
Here, $z_t^{(k)}$ denotes the last-layer hidden state at position $t$ produced by the $k$-th forward pass.
The model generates these latent-step representations autoregressively, which introduces a strict \emph{sequential dependency} across both the sequence dimension ($t$) and the latent-step dimension ($k$).

\begin{wrapfigure}{l}{0.50\textwidth}
    \vspace{-0.6\baselineskip} 
    \centering
    \includegraphics[width=\linewidth]{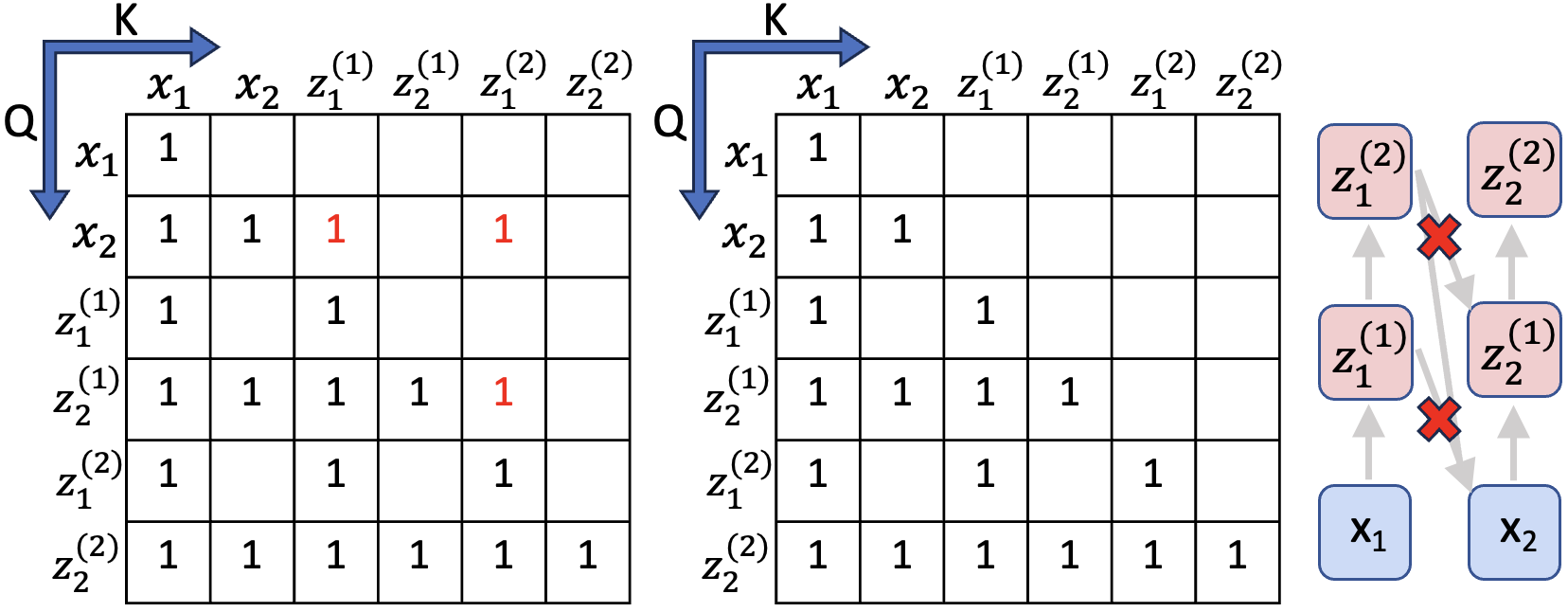}
    \captionof{figure}{
\textbf{Left:} Standard latent-CoT induces a strict sequential chain across both sequence length $L$ and latent depth $K$ (highlighted in \textcolor{red}{red}): later tokens (e.g., $x_2$) can depend on deeper latent states of earlier positions.
\textbf{Right:} Enforcing strict 2D causality with an attention mask that allows $(t_i,k_i)\!\rightarrow\!(t_j,k_j)$ only if $t_j \le t_i$ and $k_j \le k_i$ blocks cross-token dependencies on future latent steps (shown by \textcolor{red}{$\times$}), making the process sequential only in $k$ and enabling parallel computation over all $t$ at each step.
}
    \label{fig:parallel_masking_dependency}
    \vspace{-0.6\baselineskip} 
\end{wrapfigure}

\paragraph{A minimal example.}
Consider autoregressive generation for tokens $x_1$ and $x_2$ with latent steps. The hidden state $\vect{h}_{x_2}$ for $x_2$ depends on the complete resolution of all latent steps associated with $x_1$:
\begin{equation}
    \vect{h}_{x_2} = \text{LM}\!\left(x_1, z_{1}^{(1)}, \dots, z_{1}^{(K)}, x_2\right).
\end{equation}
As illustrated in~\autoref{fig:parallel_masking_dependency}(left), this means $\vect{h}_{x_2}$ cannot be computed until the entire chain $z_{1}^{(1 \dots K)}$ is finalized, since attention requires access to progressively deeper latent hidden states (highlighted by the \textcolor{red}{red} dependencies).

\paragraph{The pre-training dilemma.}
Naively training under this strictly sequential dependency incurs $O(L \times K)$ sequential operations per sequence, which becomes computationally prohibitive for long contexts (e.g., thousands of tokens). PonderLM2~\citep{zeng2025ponderlm} mitigates this by computing latent states for every token at each latent step simultaneously and then refining them with multi-step Jacobi iterations. While this enables parallelism, the iterative refinement substantially increases training compute---for example, even with a latent step budget of $K=1$, it can incur around $8\times$ the training FLOPs of the vanilla model.

\subsection{Parallel Latent Steps via Parallel Masking}
To enable parallel computation while maintaining training efficiency, we expand attention into two dimensions: \emph{sequence position} ($t$) and \emph{latent step} ($k$). We define a \textbf{Parallel Attention Mask} over 2D indices $(t,k)$ that enforces strict causality in both dimensions:
\begin{equation}
M_{(t_i, k_i), (t_j, k_j)} =
\begin{cases}
0 & \text{if } t_j \le t_i \ \land\  k_j \le k_i, \\
-\infty & \text{otherwise.}
\end{cases}
\end{equation}

This mask induces a partial order over the 2D computation graph (\autoref{fig:parallel_masking_dependency}, right): at any fixed latent step $k$, the model can compute states for the entire sequence $t \in [1,L]$ in parallel. Unlike standard autoregressive generation which is sequential in $L$, our training process effectively unrolls the latent iterations. This transforms the dependency graph from $O(L \times K)$ sequential operations to $O(K)$ sequential steps, thereby leveraging the massive parallelism of GPUs over the sequence dimension.

\subsection{Adaptive Computation via Probabilistic Halting}
\label{subsec:prob_halting}
\begin{figure*}[t]
    \centering
    \includegraphics[width=\linewidth]{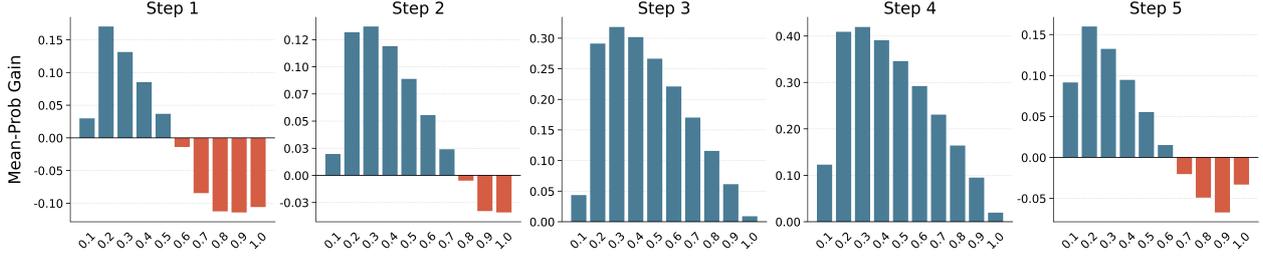}
    \caption{
    Motivation for adaptive computation and the correctness-aware adaptive loss.
    We probe a latent-CoT model without adaptive computation on 0.34B tokens.
    At each latent step $k$, we bucket tokens by their current target-token probability $p_{\mathrm{target}}^{(k)}$ (x-axis),
    and measure the average improvement in the target probability brought by the next latent step,
    $p_{\mathrm{target}}^{(k+1)} - p_{\mathrm{target}}^{(k)}$ (y-axis).
    While extra latent computation substantially improves low-confidence tokens, the gains diminish as $p_{\mathrm{target}}^{(k)}$ increases,
    and can become negative for already confident tokens (negative bars), suggesting unnecessary or even harmful computation.}
    \label{fig:adaptive_loss_motivation}
    \vskip -0.2in
\end{figure*}

While the parallel masking enables efficient unrolling over latent steps, executing many steps uniformly for all tokens is not always desirable.
Empirically, once a token is already likely correct (high $p_{\mathrm{target}}$), further latent computation yields diminishing returns and can even hurt the target probability ~\autoref{fig:adaptive_loss_motivation}.
We therefore introduce a probabilistic halting mechanism to adapt computation per token. We denote the maximum latent CoT length by $\ell_{\max}$, and implement it with $K_{\max}=\ell_{\max}+1$.

\paragraph{Router (conditional continuation).}
At each latent step $k \in \{1,\dots,K_{\max}\}$, the Router predicts the conditional probability of \textit{continuing} to the next latent step:
\begin{equation}
    g_t^{(k)} \triangleq P(\text{Continue} \mid \text{Reach step } k)
    = \sigma\!\bigl(\mathrm{Linear}(z_t^{(k)})\bigr).
\end{equation}
Accordingly, $P(\text{Stop} \mid \text{Reach step } k)=1-g_t^{(k)}$.

\paragraph{Reach probability (accumulated probability flow).}
For computation to exist at step $k$, it must have continued through all previous steps. We define the reach probability recursively:
\begin{equation}
    p_{\mathrm{reach},t}^{(1)} \triangleq 1,\qquad
    p_{\mathrm{reach},t}^{(k+1)} \triangleq p_{\mathrm{reach},t}^{(k)}\, g_t^{(k)}.
\end{equation}

\paragraph{Exit probability (halting distribution).}
The probability of halting \emph{exactly} at step $k$ is the joint probability of reaching step $k$ and deciding to stop:
\begin{equation}
    p_{\mathrm{exit},t}^{(k)} \triangleq P(K_t=k)
    = p_{\mathrm{reach},t}^{(k)}\bigl(1-g_t^{(k)}\bigr).
\end{equation}
We set $g_t^{(K_{\max})}=0$ so that the process always halts by $K_{\max}$, yielding
$p_{\mathrm{exit},t}^{(K_{\max})}=p_{\mathrm{reach},t}^{(K_{\max})}$.

\paragraph{Truncation (threshold pruning).}
To save FLOPs, we avoid executing latent steps that are unlikely to be reached. After computing $g_t^{(k)}$, we estimate the probability of reaching the next step via $p_{\mathrm{reach},t}^{(k+1)}$.
If $p_{\mathrm{reach},t}^{(k+1)} < \tau$, we prune token $t$ from the batch for all subsequent steps.
Equivalently, we define the last executed step:
\begin{equation}
    K_t^\star \triangleq \max\Bigl\{k \le K_{\max}\ \bigm|\ p_{\mathrm{reach},t}^{(k)} \ge \tau\Bigr\}.
\end{equation}

\paragraph{Expectation-based mixing with residual re-allocation.}
We compute the final token representation as the expectation of latent-step states under the (truncated) halting distribution. Since pruning truncates the unrolled process, na\"ively using $\{p_{\mathrm{exit},t}^{(k)}\}_{k\le K_t^\star}$ would discard probability mass beyond $K_t^\star$.
We therefore re-allocate the residual mass to the terminal executed state:
\begin{equation}
    \hat p_{\mathrm{exit},t}^{(k)}=
    \begin{cases}
    p_{\mathrm{exit},t}^{(k)} & \text{if } k < K_t^\star,\\[2pt]
    p_{\mathrm{reach},t}^{(K_t^\star)}
    \;=\;\displaystyle\sum_{k'=K_t^\star}^{K_{\max}} p_{\mathrm{exit},t}^{(k')}
    & \text{if } k = K_t^\star,
    \end{cases}
\end{equation}

which preserves total mass $\sum_{k=1}^{K_t^\star}\hat p_{\mathrm{exit},t}^{(k)}=1$.
Finally, the output embedding is
\begin{equation}
    z_t^{\mathrm{final}}
    = \sum_{k=1}^{K_t^\star} \hat p_{\mathrm{exit},t}^{(k)}\, z_t^{(k)}.
\end{equation}
Importantly, the Router gates $\{g_t^{(k)}\}$ affect $\hat p_{\mathrm{exit},t}^{(k)}$ (via the probability flow $p_{\mathrm{reach}}\!\rightarrow\!p_{\mathrm{exit}}$), hence gradients from the main CE loss backpropagate to the Router through the mixing weights.

\subsection{Correctness-Aware Adaptive Loss}
\label{subsec:adaptive_loss}

While probabilistic halting enables token-wise adaptive computation, we additionally need to guide the Router to stop when extra computation is unlikely to help.
As evidenced in ~\autoref{fig:adaptive_loss_motivation}, once $p_{\mathrm{target}}^{(k)}$ is high, additional latent steps bring little benefit and can even be detrimental.
We therefore propose a \emph{Correctness-Aware Adaptive Loss} that encourages early halting for tokens that are already likely correct, reducing overall FLOPs.
Let
\begin{equation}
    p_{\mathrm{target},t}^{(k)} \triangleq \bigl[\mathrm{LMHead}(z_t^{(k)})\bigr]_{y_t}
\end{equation}
denote the probability assigned to the ground-truth token $y_t$ at step $k$. We encourage the model to halt when it is already likely correct by penalizing continuation proportional to this step-wise target probability:
\begin{equation}
    \mathcal{L}_{\mathrm{adaptive}}
    = \lambda \sum_{t} \sum_{k=1}^{K_t^\star}
    g_t^{(k)} \cdot \mathrm{sg}\!\left( (p_{\mathrm{target},t}^{(k)})^\beta \right),
    \label{eq:adaptive_loss}
\end{equation}
where $\mathrm{sg}(\cdot)$ denotes stop-gradient to prevent degenerate solutions (e.g., reducing $p_{\mathrm{target}}$ solely to weaken the compute penalty), $\beta$ controls the nonlinearity of the weighting (larger $\beta$ concentrates the penalty on high-$p_{\mathrm{target}}$ steps), and $\lambda$ scales the overall strength of the adaptive term relative to $\mathcal{L}_{\mathrm{CE}}$. Intuitively, when $p_{\mathrm{target}}^{(k)}$ is high, the penalty encourages halting ($g\!\to\!0$); when the model is uncertain, the penalty is small, allowing additional latent computation.

The overall objective is:
\begin{equation}
    \mathcal{L}
    = \mathcal{L}_{\mathrm{CE}}\!\Bigl(\mathrm{LMHead}(z^{\mathrm{final}}), y\Bigr)
    + \mathcal{L}_{\mathrm{adaptive}}.
\end{equation}

\subsection{Training and Inference Workflows}
\label{subsec:workflow}

We now summarize how the above components are integrated in practice.
\begin{figure*}[t]
    \centering
    \includegraphics[width=\textwidth]{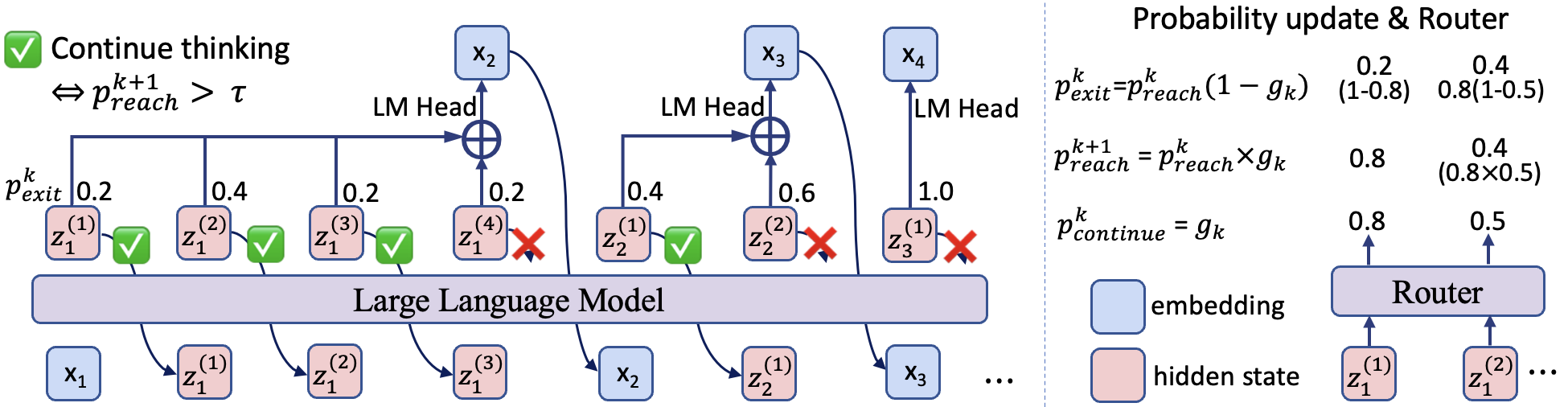}
\caption{Inference: token-wise adaptive latent CoT via reach probability.
\textbf{Left:} At decoding position $t$, the model iterates latent steps to produce $z_t^{(1)},z_t^{(2)},\ldots$ and updates the reach probability, stopping when the next-step reach probability $p_{\mathrm{reach},t}^{(k+1)}$ falls below the threshold $\tau$.
Easy tokens ($x_2,x_3$) stop earlier (shorter latent CoT), while harder tokens ($x_1$) continue longer.
The final representation is obtained by $p_{\mathrm{exit}}$-weighted mixing of the executed latent states, and is fed to the LM head for prediction.
\textbf{Right:} The Router takes each latent hidden state $z_t^{(k)}$ and outputs a gate $g_t^{(k)}$ (the conditional continuation probability), which defines the probability flow:
$p_{\mathrm{reach}}^{k+1}=p_{\mathrm{reach}}^{k} g^{k}$ and
$p_{\mathrm{exit}}^{k}=p_{\mathrm{reach}}^{k}(1-g^{k})$.}
\label{fig:inference_halting}
\vskip -0.1in
\end{figure*}

\paragraph{Training.}
\begin{figure*}[t]
    \centering
    \includegraphics[width=\textwidth]{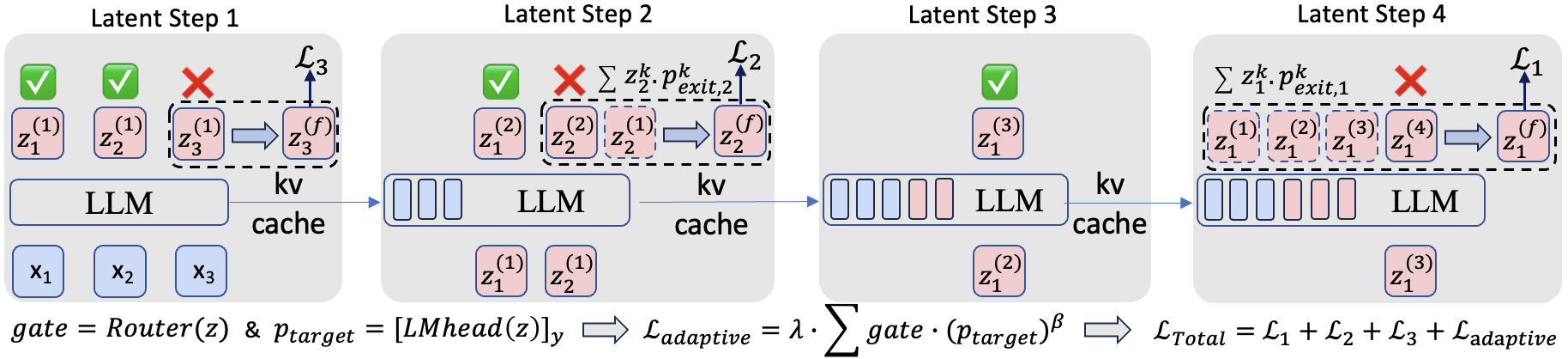}
\caption{A training example with unrolled latent steps, pruning, and KV-cache reuse. We illustrate one sequence $(x_1,x_2,x_3)$ with a latent budget $K_{\max}=4$.
With the parallel attention mask, we unroll computation along latent steps and, at each step, compute all \emph{active} tokens in parallel.
At step 1, the model produces $(z_1^{(1)},z_2^{(1)},z_3^{(1)})$ and the Router outputs gates (conditional continuation probabilities) for each token, which update $p_{\mathrm{reach}}$ and determine whether a token remains active.
In the figure, $x_1,x_2$ keep sufficient reach probability to proceed (green checks), while $x_3$ is pruned (red cross).
At step 2, only $x_1,x_2$ remain; the updated reach probability prunes $x_2$, so steps 3--4 are executed only for $x_1$.
Across latent steps, we reuse the KV cache so later steps reuse cached attention context, avoiding redundant computation.
Since pruning shrinks the active token set as $k$ increases, the overall training FLOPs decrease.
For supervision, we compute per-position LM losses ($\mathcal{L}_1,\mathcal{L}_2,\mathcal{L}_3$) on the final representation $z_t^{\mathrm{f}}$ obtained by $p_{\mathrm{exit}}$-weighted mixing of the executed latent states, and add a correctness-aware adaptive loss (proportional to $p_{\mathrm{target}}$) to penalize continuing when the model already assigns high probability to the ground-truth token, the $p_{\mathrm{exit}}$-weighted mixing and $\mathcal{L}_{\mathrm{adaptive}}$ provide learning signals to both the LM and the Router.}

\label{fig:train_workflow}
\end{figure*}
We unroll latent computation for $k=1,\ldots,K_{\max}$ under the parallel attention mask.
At each latent step, all \emph{active} token positions are processed in parallel over $t$.
To improve efficiency, we reuse the KV cache across latent steps to avoid recomputing shared attention context, and apply threshold pruning so the active token set shrinks as $k$ increases, reducing overall training FLOPs (\autoref{fig:train_workflow}).
The Router is trained jointly through (i) gradients from the main cross-entropy loss computed on the mixed representation $z^{\mathrm{final}}$ and (ii) the correctness-aware adaptive penalty that discourages unnecessary continuation.

\paragraph{Inference.}

We use the same Router to perform token-wise halting during decoding.
For each position, we iteratively compute latent states and update the reach probability, stopping when the next-step reach probability falls below a threshold $\tau$.
We then form $z^{\mathrm{final}}$ via the (truncated) $p_{\mathrm{exit}}$-weighted mixing of the executed latent states and emit the next token with the LM head (\autoref{fig:inference_halting}).

\subsection{Positional Embedding}
All latent-step states $z_t^{(k)}$ share the same positional embedding (position id) as the input token $x_t$ at position $t$, rather than using new positions. Thus, latent steps do not consume extra position ids, and \textbf{the effective context window stays unchanged} from the vanilla model.

\section{Experiments}
\label{sec:experiments}

\subsection{Setup}
\label{subsec:main_results}

\paragraph{Training Setup and Baselines.}
We train all models from scratch using the LLaMA architecture~\citep{dubey2024llama}. Specifically, we pretrain two model sizes (410M and 1.4B parameters) for 26B tokens on the Pile over 50{,}000 optimization steps, following the protocol of~\cite{zeng2025ponderlm}.
To ensure a controlled comparison, we train a suite of baselines under an \emph{identical pretraining protocol}---matching the training data, optimization schedule, and hyperparameters---including vanilla LLaMA models as well as competitive compute-scaling and recurrent variants: LoopedLM~\citep{giannou2023looped}, PausedLM~\citep{goyal2023think}, PonderLM~\citep{zeng2025pretraining}, PonderLM2~\citep{zeng2025ponderlm}, and the adaptive Mixture-of-Recurrence (MoR) method~\citep{bae2025mixture}.
\begin{figure*}[t] 
    \centering
    \includegraphics[width=\textwidth]{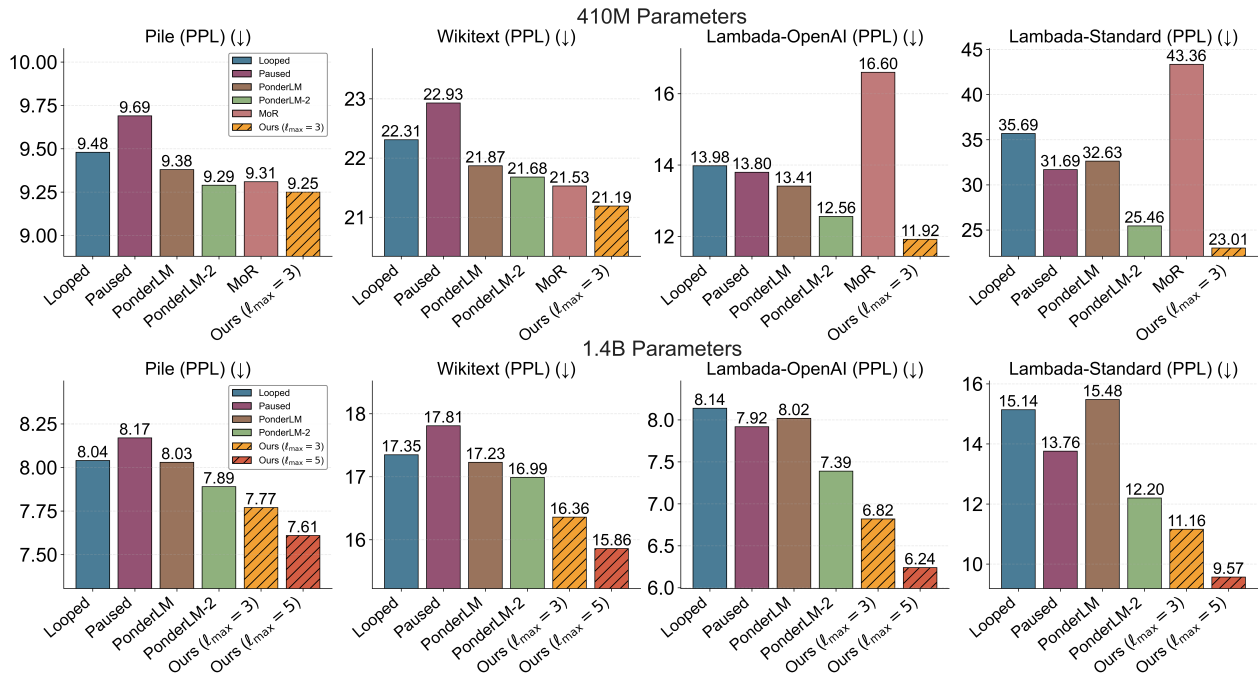}
    \caption{
\textbf{Language modeling perplexity (PPL $\downarrow$).}
Validation perplexity  on The Pile, WikiText, and LAMBADA (OpenAI and standard splits).
Our method consistently achieves the lowest perplexity across all datasets while using fewer training FLOPs.
}
\label{fig:perplexity_results}
\vskip -0.2in
\end{figure*}
\paragraph{Evaluation.}
We evaluate language modeling quality using validation perplexity on The Pile~\citep{gao2020pile}, WikiText~\citep{merity2016pointer}, and LAMBADA~\citep{paperno2016lambada} (both the OpenAI and standard splits).
To assess downstream generalization, we follow the evaluation protocol of~\cite{gu2024mamba} and report 0-shot and 5-shot accuracy on a broad benchmark suite, including LAMBADA~\citep{paperno2016lambada}, SciQ~\citep{welbl2017crowdsourcing}, HellaSwag~\citep{zellers2019hellaswag}, PIQA~\citep{bisk2020piqa}, WinoGrande~\citep{sakaguchi2021winogrande}, ARC-Easy and ARC-Challenge~\citep{clark2018think}, and RACE~\citep{lai2017race}.

\subsection{Main Results}
\paragraph{Language Modeling Ability.} As shown in~\autoref{fig:perplexity_results}, our method consistently achieves the lowest perplexity across all datasets while using the least training FLOPs.
Specifically, our LLaMA-1.4B model with max latent CoT length $\ell_{\max} = 3$ outperforms the strongest baseline (PonderLM-2) while requiring less than half of the training compute (7.47 vs.\ 17.47 $\times 10^{20}$ FLOPs). Moreover, increasing $\ell_{\max}$ to 5 further improves performance.
\paragraph{Downstream Tasks.}
 Across both zero-shot and five-shot settings and under all compute budgets, our method achieves the best overall average accuracy (\autoref{tab:downstream_details}). Notably, \textbf{our LLaMA-410M} with $\ell_{\max} = 3$ \emph{outperforms} the compute-comparable vanilla \textbf{LLaMA-1.4B} baseline in average accuracy for both zero-shot and five-shot, indicating that our approach can be even more effective than simply scaling up the parameter count under a matched compute budget.

\definecolor{lightergray}{gray}{0.9}

\begin{table*}[t!]
    \centering
    \caption{
Zero-shot and five-shot accuracy (\%) on downstream tasks. FLOPs denote  total pretraining compute (in units of $\times 10^{20}$), estimated following~\cite{kaplan2020scaling}.
Rows are grouped by compute budget. $^\dagger$ denotes a vanilla baseline with comparable compute.
}
\label{tab:downstream_details}

    \small 
    \setlength{\tabcolsep}{3.2pt}      
    \renewcommand{\arraystretch}{0.95} 

    \resizebox{\textwidth}{!}{%
    \begin{tabular}{lccccccccccc}
    \toprule
    \textbf{Model} & \textbf{FLOPs} & \makecell{\textbf{Lambada}\\\textbf{(OpenAI)}} & \makecell{\textbf{Lambada}\\\textbf{(Std)}} & \makecell{\textbf{ARC}\\\textbf{-C}} & \makecell{\textbf{ARC}\\\textbf{-E}} & \makecell{\textbf{Wino}\\\textbf{Grande}} & \textbf{PIQA} & \makecell{\textbf{Hella}\\\textbf{Swag}} & \textbf{SciQ} & \textbf{RACE} & \makecell{\textbf{Avg.}\\\textbf{Acc.}} \\
    \midrule\midrule

    \multicolumn{12}{c}{\textit{Zero-shot Performance}} \\
    \midrule\midrule

    Looped LLaMA-410M (4 loops) & 2.56 & 47.1 & 34.9 & 22.6 & 50.9 & 51.2 & 65.5 & 32.6 & 79.5 & 30.4 & 46.1 \\
    Pause LLaMA-410M (3 pauses) & 2.56 & 46.3 & 35.3 & 20.9 & 49.9 & 51.7 & 64.7 & 31.7 & 80.1 & 30.4 & 45.7 \\
    PonderLM-410M (3 steps)     & 2.56 & 48.9 & 36.9 & 21.2 & 50.4 & \textbf{54.1} & \textbf{66.3} & 33.3 & 80.6 & 30.4 & 46.9 \\
    PonderLM2-410M (1 step)    & 5.12 & 48.7 & 37.7 & 21.0 & 50.6 & 51.3 & 65.1 & 33.0 & 79.8 & 30.8 & 46.4 \\
    MoR-410M (4 Recursions)     & 2.27 & 43.1 & 32.0 & 19.8 & 49.5 & 51.7 & 64.9 & 31.1 & 80.8 & 30.9 & 44.9 \\
    LLaMA-1.4B (train from scratch)$^\dagger$ & 2.18 & \textbf{50.6} & 37.1 & 20.9 & \textbf{52.2} & 53.0 & 65.6 & 33.6 & \textbf{84.5} & 31.8 & 47.7 \\
    \rowcolor{lightergray}
    \textbf{\boldmath Our LLaMA-410M ($\ell_{\max} = 3$)} 
    & 2.27 & 49.3 & \textbf{40.2} & \textbf{23.6} & 50.8 & 53.7 & 65.9 & \textbf{33.9} & 82.2 & \textbf{33.0} & \textbf{48.1} \\
    \midrule

    Looped LLaMA-1.4B (4 loops) & 8.74 & 55.8 & 45.6 & 23.2 & 55.2 & 54.1 & 68.9 & 37.6 & 84.9 & 33.0 & 50.9 \\
    Pause LLaMA-1.4B (3 pauses) & 8.74 & 56.2 & 46.5 & 24.2 & 54.0 & 55.3 & 68.8 & 36.7 & 85.4 & 32.3 & 51.0 \\
    PonderLM-1.4B (3 steps)     & 8.74 & 56.7 & 45.4 & 23.8 & 56.3 & 55.6 & 68.3 & 37.8 & 86.3 & 33.0 & 51.5 \\
    \rowcolor{lightergray}
    \textbf{\boldmath Our LLaMA-1.4B ($\ell_{\max} = 3$)}
    & 7.47 & \textbf{58.8} & \textbf{49.4} & \textbf{25.8} & \textbf{57.1} & \textbf{57.4} & \textbf{69.2} & \textbf{39.3} & \textbf{87.2} & \textbf{33.4} & \textbf{53.1} \\
    \midrule

    PonderLM-1.4B (4 steps)     & 10.92 & 56.2 & 45.8 & 24.9 & 54.5 & 54.7 & 69.0 & 38.1 & 84.3 & \textbf{35.4} & 51.4 \\
    PonderLM2-1.4B (1 step)    & 17.47 & 58.1 & 48.2 & 25.2 & 58.0 & 53.9 & \textbf{70.7} & 38.6 & 85.9 & 32.4 & 52.3 \\
    \rowcolor{lightergray}
    \textbf{\boldmath Our LLaMA-1.4B ($\ell_{\max} = 5$)}
    & 10.84 & \textbf{61.0} & \textbf{52.7} & \textbf{25.9} & \textbf{59.2} & \textbf{57.6} & 70.4 & \textbf{40.9} & \textbf{87.1} & 35.3 & \textbf{54.5} \\
    \midrule\midrule

    \multicolumn{12}{c}{\textit{Five-shot Performance}} \\
    \midrule\midrule

    Looped LLaMA-410M (4 loops) & 2.56 & 42.0 & 33.6 & 22.7 & 52.2 & 50.9 & 66.2 & 33.0 & 86.9 & 30.5 & 46.4 \\
    Pause LLaMA-410M (3 pauses) & 2.56 & 40.1 & 32.0 & 22.4 & 50.3 & \textbf{52.8} & 65.6 & 31.7 & 84.8 & 30.8 & 45.6 \\
    PonderLM-410M (3 steps)     & 2.56 & 42.3 & \textbf{36.0} & 23.5 & 52.6 & \textbf{52.8} & 66.9 & 33.4 & 87.5 & 31.5 & 47.4 \\
    PonderLM2-410M (1 step)    & 5.12 & 43.5 & 34.4 & 21.7 & 53.3 & 51.8 & 66.3 & 33.2 & 88.5 & 31.2 & 47.1 \\
    MoR-410M (4 Recursions)     & 2.27 & 40.3 & 32.3 & 22.3 & 52.3 & 50.0 & 66.8 & 33.2 & 88.7 & 31.1 & 46.3 \\
    LLaMA-1.4B (train from scratch)$^\dagger$ & 2.18 & \textbf{45.1} & 34.7 & 22.3 & 53.5 & 50.9 & 66.1 & 33.6 & 89.3 & 31.6 & 47.5 \\
    \rowcolor{lightergray}
    \textbf{\boldmath Our LLaMA--410M ($\ell_{\max} = 3$)}
    & 2.27 & 42.7 & 35.7 & \textbf{23.6} & \textbf{53.8} & 52.3 & \textbf{67.3} & \textbf{33.7} & \textbf{89.8} & \textbf{32.0} & \textbf{47.9} \\
    \midrule

    Looped LLaMA-1.4B (4 loops) & 8.74 & 48.0 & 42.8 & 25.0 & 58.5 & 54.8 & \textbf{70.4} & 37.6 & 89.2 & 28.5 & 50.5 \\
    Pause LLaMA-1.4B (3 pauses) & 8.74 & 48.8 & 41.0 & 24.1 & 56.6 & 54.1 & 69.3 & 36.6 & 90.7 & 32.6 & 50.4 \\
    PonderLM-1.4B (3 steps)     & 8.74 & 49.5 & \textbf{42.6} & 25.4 & 58.8 & 54.3 & 69.2 & 37.9 & 91.2 & \textbf{34.7} & 51.5 \\
    \rowcolor{lightergray}
    \textbf{\boldmath Our LLaMA-1.4B ($\ell_{\max} = 3$)}
    & 7.47 & \textbf{51.7} & 41.7 & \textbf{27.7} & \textbf{60.4} & \textbf{56.0} & \textbf{70.4} & \textbf{39.6} & \textbf{92.4} & 32.7 & \textbf{52.5} \\
    \midrule

    PonderLM-1.4B (4 steps)     & 10.92 & 50.3 & 44.5 & 26.0 & 58.6 & 55.8 & 69.9 & 38.4 & \textbf{92.4} & 33.1 & 52.1 \\
    PonderLM2-1.4B (1 step)    & 17.47 & 49.8 & 45.6 & \textbf{27.7} & 59.6 & 56.3 & 69.8 & 38.7 & 91.3 & 28.0 & 51.9 \\
    \rowcolor{lightergray}
    \textbf{\boldmath Our LLaMA-1.4B ($\ell_{\max} = 5$)}
    & 10.84 & \textbf{54.8} & \textbf{51.0} & 27.5 & \textbf{62.5} & \textbf{56.8} & \textbf{71.4} & \textbf{41.2} & 91.4 & \textbf{35.5} & \textbf{54.7} \\
    \bottomrule
    \end{tabular}
    }%
\vskip -0.1in
\end{table*}

\subsection{Iso-data and Iso-FLOP Comparison}
\label{subsec:IsoFlop Analysis}
\begin{figure}[t]
    \centering
    \begin{minipage}[t]{\colfigwidth}
        \vspace{0pt}
        \centering
        \includegraphics[width=\linewidth]{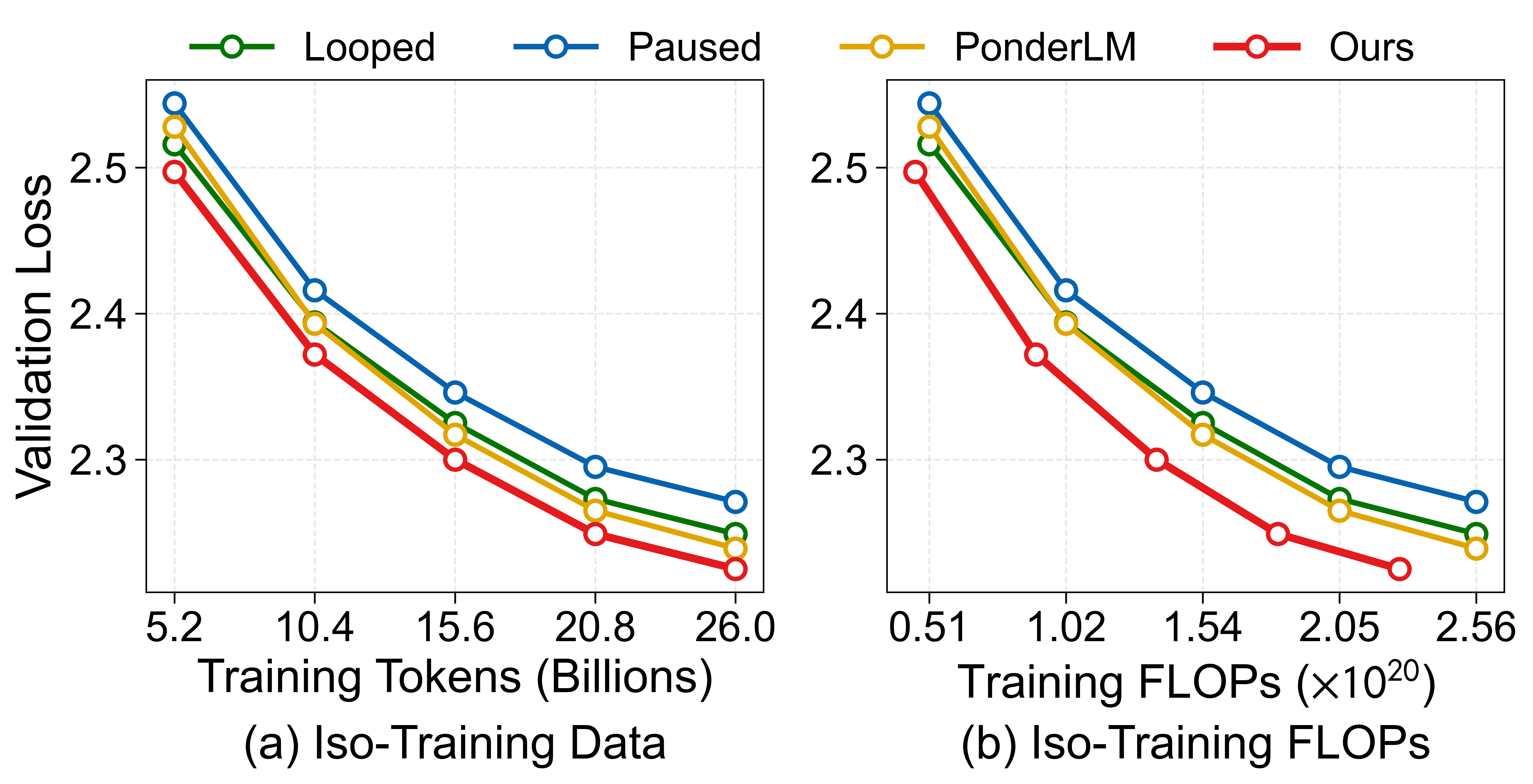}
        \captionsetup{type=figure,skip=2pt,justification=raggedright,singlelinecheck=false}
        \captionof{figure}{Validation loss versus token and compute. Our method consistently achieves the lowest validation loss across budgets.}
        \label{fig:isoflop}
    \end{minipage}\hfill
    \begin{minipage}[t]{\colfigwidth}
        \vspace{0pt}
        \centering
        \includegraphics[width=\linewidth]{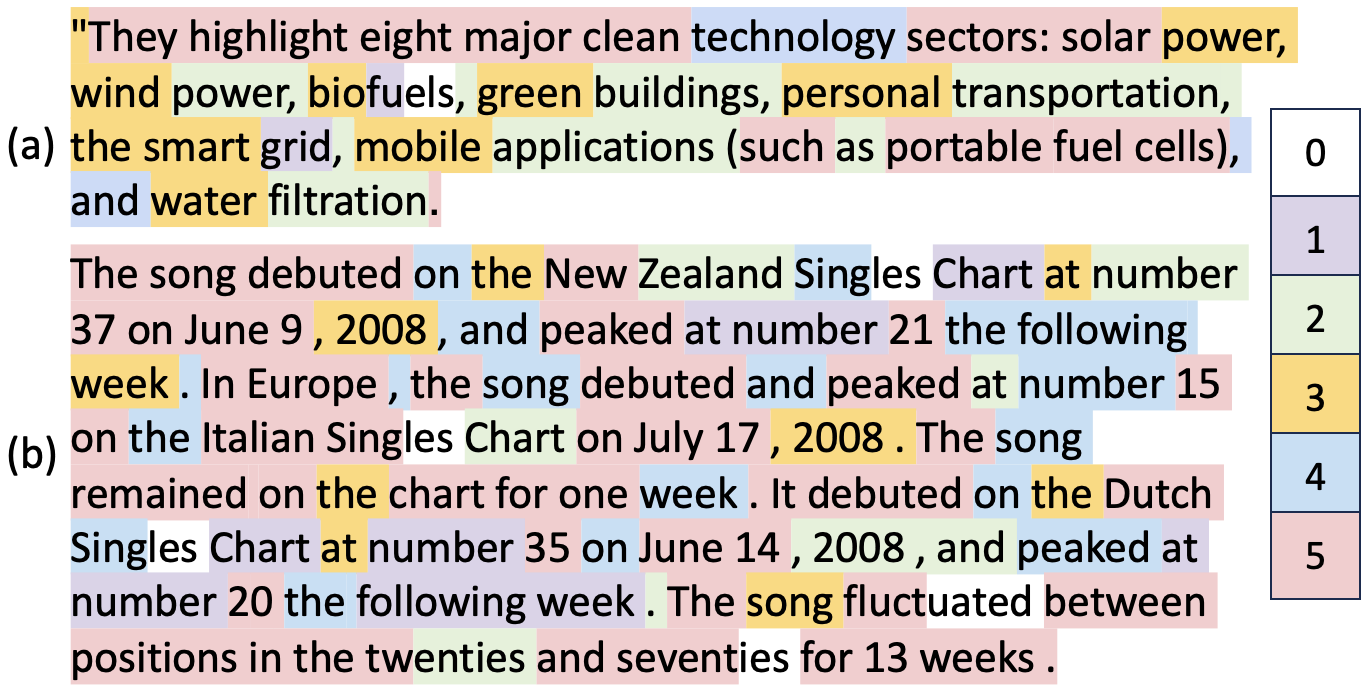}
        \captionsetup{type=figure,skip=2pt,justification=raggedright,singlelinecheck=false}
        \captionof{figure}{
\textbf{Case studies of our adaptive latent cot model.}
Two examples from the 1.4B model with $l_{\max}=5$, showing per-token executed latent steps (color-coded).
}
        \label{fig:case_study}
    \end{minipage}
    \vskip -0.15in
\end{figure}
In~\autoref{fig:isoflop}, we compare four 410M-parameter models (LoopedLM, PausedLM, PonderLM, and ours) and report their validation loss under two alignment settings: (i) \emph{iso-data}, where checkpoints are matched by the same token/step budget, and (ii) \emph{iso-FLOPs}, where checkpoints are matched by the same estimated total training compute.
Across both settings and all budgets, our method consistently achieves the lowest validation loss.

\FloatBarrier
\subsection{Analysis of Token-Level Adaptive Latent CoT}
\label{subsec:adaptive_analysis}
We analyze \emph{when} the model decides to halt or continue latent CoT, and how these decisions allocate computation across tokens.
We use our previously trained LLaMA-1.4B models with max latent CoT length $\ell_{\max}\in\{3,5\}$.
Specifically, we study:
(i) the relationship between token-wise latent CoT length and $p_{\text{target}}$ (the probability assigned to the ground-truth next token),
(ii) whether the model generates shorter latent CoT for easy tokens and longer latent CoT for difficult tokens, and
(iii) qualitative case studies illustrating where longer latent CoT is triggered in real sequences.

\subsubsection{Token-wise Latent CoT Length vs.\ $p_{\text{target}}$}
We examine how $p_{\text{target}}$ varies across tokens that execute different latent CoT lengths.
The results in~\autoref{fig:difficulty_scaling} exhibit a clear monotonic trend: tokens with longer latent CoT tend to have lower $p_{\text{target}}$, whereas tokens with higher $p_{\text{target}}$ typically halt earlier.
This observation aligns with our correctness-aware design: pruning is encouraged when the ground-truth token already receives a high probability, while ambiguous tokens are allocated more latent computation.

\subsubsection{Token Difficulty vs.\ Latent CoT Length}
\begin{figure}[!hb]
    \centering
    \begin{minipage}[t]{\colfigwidth}
        \vspace{0pt}
        \centering
        \includegraphics[width=0.98\linewidth]{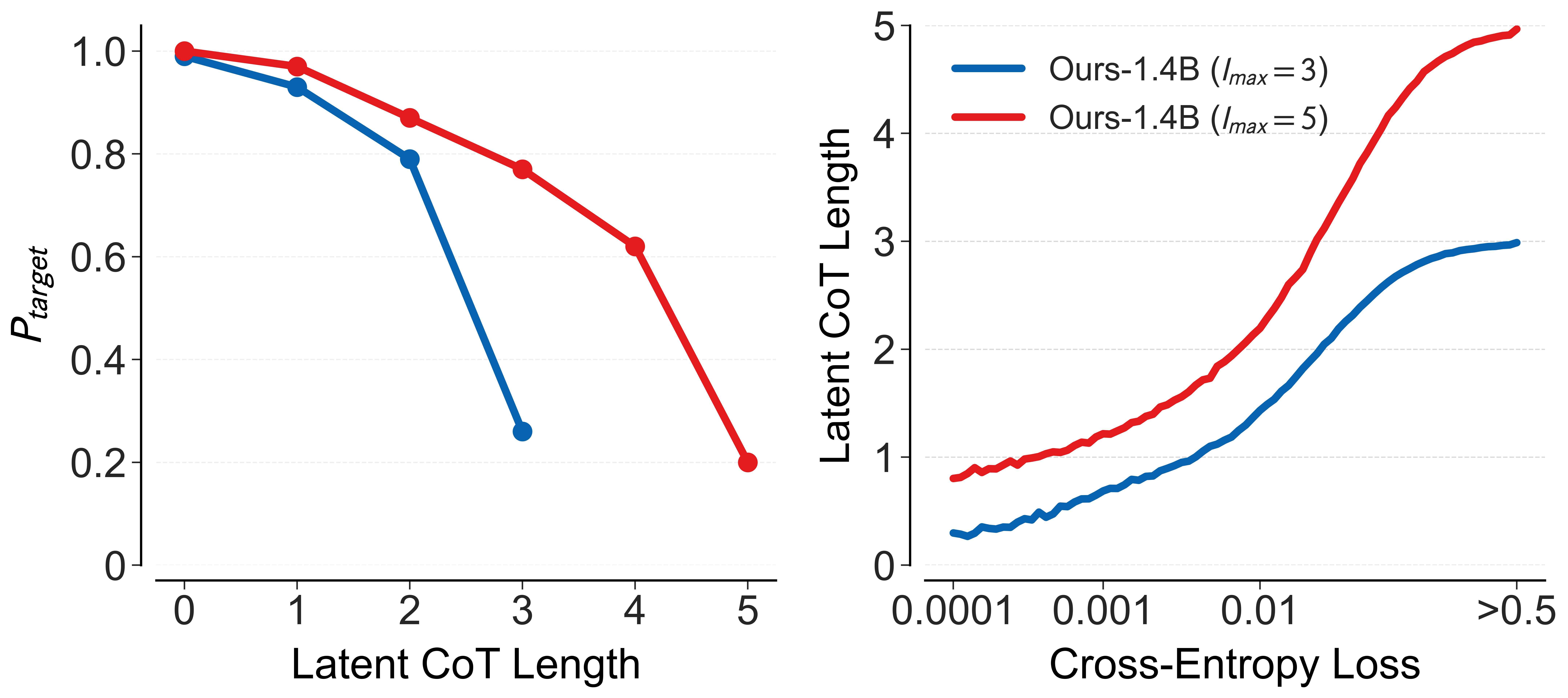}
    \end{minipage}\hfill
    \begin{minipage}[t]{\colfigwidth}
        \vspace{0pt}
        \captionsetup{
            type=figure,
            skip=2pt,
            justification=raggedright,
            singlelinecheck=false
        }
        \captionof{figure}{
\textbf{Left:} $p_{\text{target}}$ monotonically decreases with latent CoT length: tokens that generate longer latent CoT have lower $p_{\text{target}}$.
\textbf{Right:} We bucket tokens by token-wise cross-entropy (log-scale) and report average latent COT length for 1.4B models with $l_{\max}\in\{3,5\}$. Easy tokens use $\sim$0--1 steps, while harder tokens receive more computation.
}
        \label{fig:difficulty_scaling}
    \end{minipage}
    \vskip -0.2in
\end{figure}

On WikiText~\citep{merity2016pointer}, we use token-wise cross-entropy as a difficulty score and bucket tokens accordingly.
As shown in \autoref{fig:difficulty_scaling}, latent CoT length increases monotonically with difficulty: easy (low-loss) tokens execute $\sim$0--1 latent steps on average, while difficult (high-loss) tokens trigger deeper computation and approach the maximum length.
This suggests the model learns a sensible compute-allocation rule, using minimal latent CoT for easy tokens and longer latent CoT for difficult ones.

\subsubsection{Case Studies}
Finally, we visualize where the model chooses to generate long latent CoT in real sequences.~\autoref{fig:case_study} shows two examples from the 1.4B model with max latent CoT length $\ell_{\max}=5$, where each token is colored by the number of executed latent steps.
The router assigns minimal computation to highly predictable spans (e.g., common function words and locally determined continuations), while allocating longer latent CoT to information-dense tokens such as entities, numbers, and key content words.

\FloatBarrier
\section{Ablation Studies}
\label{subsec:ablation}
We run controlled ablations on 70M-parameter Llama models to isolate key design choices in our adaptive latent CoT.

\subsection{Effect of Router Architecture.}
We compare five router designs with the backbone and training setup fixed (\autoref{tab:router_ablation}).
\textit{No-Adaptive} disables routing (no pruning).
\textit{Linear}/\textit{MLP} share one router across latent steps, while \textit{Multi-Linear}/\textit{Multi-MLP} use step-specific routers (no sharing). As shown in \autoref{tab:router_ablation}, adaptive routing improves efficiency while also improving performance.
The shared \textit{Linear} router achieves the lowest CE loss, and we use it as the default.

\begin{table}[t]
    \centering
\caption{
Effect of router architecture ($l_{\max}=5$).
\textit{No-Adaptive} disables routing (no pruning).
\textit{Linear}/\textit{MLP} share one router across latent steps, while \textit{Multi-Linear}/\textit{Multi-MLP} use step-specific routers.
We report CE loss (lower is better) and prune ratio (higher saves more compute).
Notably, \textbf{shared Linear outperforms No-Adaptive in CE loss} while pruning $\sim$12\% of steps.
}
\label{tab:router_ablation}

    \vskip 0.1in
    \begin{small}
    {\scshape
    \setlength{\tabcolsep}{5pt}
    \begin{tabular}{l c c c c}
        \toprule
        \textbf{Router} & $\lambda$ & $\beta$ & \textbf{CE} $\downarrow$ & \textbf{Prune Ratio (\%)} $\uparrow$ \\
        \midrule
        No-Adaptive   & --  & --  & 2.671 & -- \\
        MLP           & 0.4 & 10  & 2.675 & 13.4 \\
        Linear        & 0.4 & 10  & \textbf{2.670} & 12.4 \\
        Multi-MLP     & 0.4 & 10  & 2.672 & 13.8 \\
        Multi-Linear  & 0.4 & 10  & 2.672 & 11.8 \\
        \bottomrule
    \end{tabular}
    }
    \end{small}
    \vskip -0.1in
\end{table}

\subsection{Effect of $\lambda$ and $\beta$ in the Adaptive Loss.}
\begin{figure*}[!hb]
    \centering
    \includegraphics[width=\linewidth]{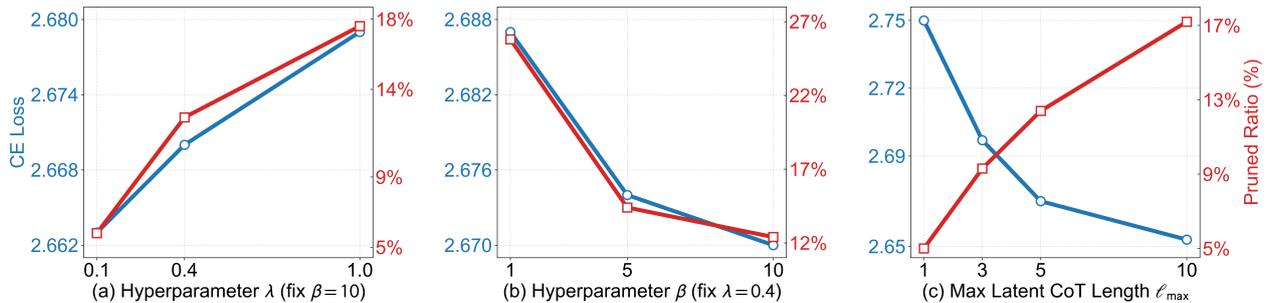}
    \caption{
\textbf{Ablations on loss hyperparameters and max latent cot length $l_{\max}$.}
Blue curves (left axis) report CE loss (lower is better), and red curves (right axis) report prune ratio (higher indicates more pruning), defined as $1-\frac{\mathbb{E}[\ell]}{l_{\max}}$, where $\ell$ denotes the executed latent CoT length per token.
(a) Vary $\lambda$ with $\beta{=}10$ and $l_{\max}{=}5$.
(b) Vary $\beta$ with $\lambda{=}0.4$ and $l_{\max}{=}5$.
(c) Vary $l_{\max}$ with $\lambda{=}0.4$ and $\beta{=}10$.
}
\label{fig:ablation_tradeoffs}
\end{figure*}

We ablate the two key hyperparameters in our correctness-aware adaptive loss (\autoref{eq:adaptive_loss}): the compute weight $\lambda$ and the exponent $\beta$ on $p_{\text{target}}$.
All runs use a shared linear router with $l_{\max}{=}5$; when varying $\lambda$ we fix $\beta{=}10$, and when varying $\beta$ we fix $\lambda{=}0.4$.
As shown in \autoref{fig:ablation_tradeoffs}(a), larger $\lambda$ increases the prune ratio but slightly worsens CE loss, reflecting a quality--efficiency trade-off; meanwhile \autoref{fig:ablation_tradeoffs}(b) shows that larger $\beta$ improves CE loss but reduces pruning, indicating more conservative early exits focused on highly confident tokens.
We adopt $\lambda{=}0.4$ and $\beta{=}10$ in our experiments.

\subsection{Effect of max latent CoT length $l_{\max}$.}
We vary $l_{\max}$ and evaluate its impact on CE loss and pruning.
As shown in \autoref{fig:ablation_tradeoffs}(c), larger $l_{\max}$ consistently reduces CE loss, suggesting extra latent cot improves modeling quality.
Meanwhile, the prune ratio also increases, indicating the router uses the larger budget to skip more steps on easy tokens while allocating computation to difficult ones.

\section{Conclusion}
We introduced Adaptive Latent COT, a one-stage pretraining framework that increases per-token computation by generating variable-length latent COT before emitting each token. Experiments on LLaMA backbones show consistent improvements in language-modeling perplexity and downstream tasks compared to strong compute-scaling baselines.

\section*{Acknowledgement}
This work is sponsored by the National Natural Science Foundation of China (NSFC) grant (No. 62576211) and the National Key Research and Development Program of China (No. 2023ZD0121402).





\newpage
\bibliography{example_paper}
\bibliographystyle{icml2026}

\newpage
\appendix
\onecolumn


\end{document}